%% file: paper.tex
\documentclass[10pt,twocolumn,letterpaper]{article}

\usepackage[pagenumbers]{cvpr} % To force page numbers, e.g. for an arXiv version

\usepackage{graphicx}
\usepackage{amsmath}
\usepackage{amssymb}
\usepackage{booktabs}
\usepackage{siunitx}
\usepackage{tabularx}

\usepackage[accsupp]{axessibility}  % Improves PDF readability for those with disabilities.

\usepackage{multibib}
\newcites{latex}{References}

\usepackage[pagebackref,breaklinks,colorlinks]{hyperref}

\usepackage[capitalize]{cleveref}
\crefname{section}{Sec.}{Secs.}
\Crefname{section}{Section}{Sections}
\Crefname{table}{Table}{Tables}
\crefname{table}{Tab.}{Tabs.}

\def\cvprPaperID{14} % *** Enter the CVPR Paper ID here
\def\confName{CVPR}
\def\confYear{2023}

\newcommand{\tradeoff}{cost-accuracy trade-off}

\newcommand{\pool}{\psi }
\newcommand{\pooladaptive}{{\psi}_{a} }
\newcommand{\R}{\mathbb{R}}

\newcommand{\myparagraph}[1]{\smallskip\noindent\textbf{#1}\hspace{0.5em}}
\newcommand{\myparagraphnospace}[1]{\noindent\textbf{#1}\hspace{0.5em}}

\newcommand\blfootnote[1]{%
  \begingroup
  \renewcommand\thefootnote{}\footnote{#1}%
  \addtocounter{footnote}{-1}%
  \endgroup
}

\usepackage{fancyhdr}
\usepackage{setspace}

\fancypagestyle{firststyle}
{

    \fancyhf{}
    \lfoot{{\footnotesize\begin{spacing}{.5}\parbox{\linewidth}{\vspace{2.5em}
    To appear in Proceedings of the \emph{IEEE/CVF Conference on Computer Vision and Pattern Recognition Workshops (CVPRW)}, The 6$^\text{th}$ Efficient Deep Learning for Computer Vision (ECV) Workshop, 2023.%
    \\\hrule\vspace{\baselineskip}
    \copyright~2023 IEEE. Personal use of this material is permitted. Permission from IEEE must be obtained for all other uses, in any current or future media, including reprinting/republishing this material for advertising or promotional purposes, creating new collective works, for resale or redistribution to servers or lists, or reuse of any copyrighted component of this work in other works. 
    }\end{spacing}}}
}

\begin{document}

%\fancyfoot[LE,RO]{\thepage}
%\fancyfoot[LO,CE]{From: K. Grant}
%\fancyfoot[CO,RE]{To: Dean A. Smith}

\title{Content-Adaptive Downsampling in Convolutional Neural Networks}

\author{%
  Robin Hesse\textsuperscript{1}
  % examples of more authors
   \and
   Simone Schaub-Meyer\textsuperscript{1,2}
   \and
   Stefan Roth\textsuperscript{1,2} \\
   \and
   \textsuperscript{1}Department of Computer Science, TU Darmstadt \quad \textsuperscript{2}hessian.AI\\
   \texttt{\{robin.hesse, simone.schaub, stefan.roth\}@visinf.tu-darmstadt.de}
}

\maketitle

%%%%%%%%% ABSTRACT
\begin{abstract}
\input{sections/abstract}
\end{abstract}
\thispagestyle{firststyle}
\section{Introduction}
\input{sections/introduction}

\section{Related Work}
\input{sections/related_work}
\section{Content-adaptive Downsampling in CNNs}
\label{sec:methods}
\input{sections/method}

\section{Experiments}\label{sec:experiments}
\input{sections/experiments}

\section{Conclusion and Discussion}\label{sec:conclusion}
\input{sections/conclusion}

\paragraph{Acknowledgements.}
This project has received funding from the
European Research Council (ERC) under the European Union’s Horizon 2020 research and innovation programme (grant agreement No.~866008).
The project has also been supported in part by the State of Hesse through the cluster project ``The Adaptive Mind (TAM)'' and the Hessian research priority programme LOEWE within the project ``WhiteBox''.

%%%%%%%%% REFERENCES
{\small
\bibliographystyle{ieee_fullname}
\bibliography{bibtex/short,bibtex/papers,bibtex/external,bibtex/local}
}

\clearpage
\newpage

\appendix
\pagenumbering{roman}

\input{sections/supplementary}

{\small

\bibliographystylelatex{ieee_fullname}
\bibliographylatex{bibtex/short,bibtex/papers,bibtex/external,bibtex/local}
}

\end{document}

%% file: sections/abstract.tex
Many convolutional neural networks (CNNs) rely on progressive downsampling of their feature maps to increase the network's receptive field and decrease computational cost. 
However, this comes at the price of losing granularity in the feature maps, limiting the ability to correctly understand images or recover fine detail in dense prediction tasks. 
To address this, common practice is to replace the last few downsampling operations in a CNN with dilated convolutions, allowing to retain the feature map resolution without reducing the receptive field, albeit increasing the computational cost. 
This allows to trade off predictive performance against cost, depending on the output feature resolution.
By either regularly downsampling or \emph{not} downsampling the \emph{entire} feature map, existing work implicitly treats all regions of the input image and subsequent feature maps as equally important, which generally does not hold. 
We propose an \emph{adaptive} downsampling scheme that generalizes the above idea by allowing to process informative regions at a higher resolution than less informative ones.
In a variety of experiments, we demonstrate the versatility of our adaptive downsampling strategy and empirically show that it improves the \tradeoff\ of various established CNNs.\blfootnote{Code available at \url{https://github.com/visinf/cad/}.}

%% file: sections/introduction.tex
\begin{figure}
  \centering
  \includegraphics[width=.95\linewidth]{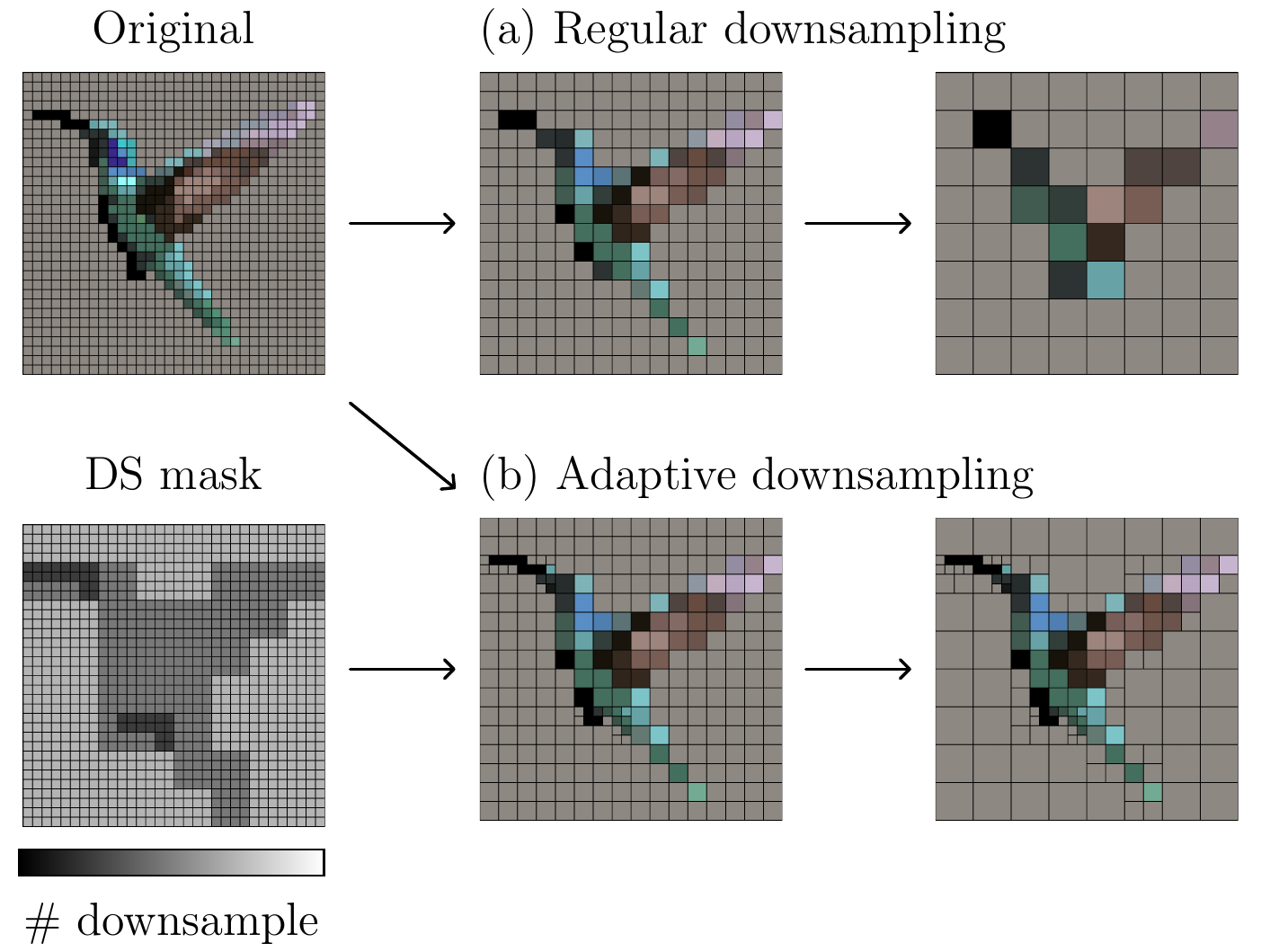}
  \vspace{-5pt}
  \caption[Figure 2]{\emph{Illustration of our content-adaptive method compared to regular downsampling.}
  \emph{(a)}~In regular downsampling, every second pixel is sampled (repeated twice here). 
  Note how the result is almost incognizable and important detail like the beak, claws, and tail of the bird is lost.
  \emph{(b)}~In our adaptive downsampling, a precomputed downsampling mask defines the number of downsampling operations applied to each pixel. 
  The resulting representation (zoom in to avoid aliasing) contains information of locally varying resolution. 
  This allows to semi-continuously adjust the representation size to keep more detail where it matters. }\label{fig:fig1_right}
\end{figure}

When humans are exposed to a complex task, they focus on relevant aspects to make optimal use of the brain's limited capacities~\cite{Bruckmaier:2020:ACL}. 
For instance, when categorizing the bird's species in \cref{fig:fig1_right}, 
our brain allocates significantly more computational resources for the bird than for the background.
While naturally occurring in humans, this adaptive allocation of resources is not utilized in most of today's deep learning architectures.
In this work, we propose an adaptive downsampling scheme for Convolutional Neural Networks (CNNs)~\cite{LeCun:1989:BAH} that mimics above adaptive allocation of resources by \emph{simultaneously} processing different regions of an image or feature map \emph{at different resolutions}.

Many CNNs used as backbones in computer vision rely on a progressive application of pooling or strided convolution~\cite{He:2016:DRL,Krizhevsky:2012:INC,Simonyan:2015:VDC} to increase the network's receptive field and decrease computational cost~\cite{Yu:2017:DRN}.
However, this progressive, regular downsampling comes at the price of losing fine detail in the feature maps, limiting the network's ability to correctly understand images~\cite{Yu:2017:DRN} or recover fine detail in dense prediction tasks~\cite{Chen:2018:ECA}, such as at object boundaries.
To approach this problem, Yu \etal~\cite{Yu:2017:DRN} replace the last few downsampling operations in a CNN with dilated convolutions~\cite{Yu:2016:MSC}. Although increasing the computational cost, this allows to keep the resolution of feature maps without reducing the network's receptive field. 
However, by regularly downsampling, respectively \emph{not} downsampling~\cite{Yu:2017:DRN}, \emph{entire} feature maps, most existing CNNs implicitly assume that all regions of the input image and subsequent feature maps are equally important, which generally does not hold~\cite{Jin:2022:LDS, Marin:2019:ESL, Recasens:2018:LZS} as, \eg, also shown by locally salient attribution maps~\cite{Simonyan:2014:DIC, Sundararajan:2017:AAD, Hesse:2021:FAA}.

In contrast, our \emph{locally adaptive} downsampling scheme makes a more realistic assumption and can be considered as a generalization of both regular downsampling in CNNs and dilated convolutions~\cite{Yu:2017:DRN}.
By processing task-relevant regions at a higher resolution than unimportant ones, we can maintain the fine granularity of important regions while efficiently processing unimportant regions at a lower resolution, ultimately leading to an improved trade-off between the accuracy and computational expense of various popular CNNs.
Contrary to existing adaptive downsampling methods~\cite{Jin:2022:LDS, Marin:2019:ESL} that adaptively downsample the input image \emph{before} passing it through a CNN, our approach allows for adaptive downsampling of feature maps \emph{within} a CNN.

Specifically, we make the following contributions: \emph{(1)} We present a novel adaptive downsampling scheme for CNNs, allowing to process different feature map regions at different resolutions.
\emph{(2)} As different amounts of computational resources are being allocated for different regions, our method can be considered as a novel realization of ``focus'' in CNNs. This is fundamentally different from existing attention mechanisms, \eg, in transformers~\cite{Dosovitskiy:2021:IWW}, where attention is implemented via adaptive weighting of input features. As a consequence, an uninformative black image and an informative natural image would still require the same inference time in a transformer, while the black image would be processed faster in our proposed adaptive downsampling.
\emph{(3)} We provide an accompanying modified instantiation of submanifold sparse convolution~\cite{Graham:2018:3DS} that allows to efficiently perform standard convolution on multi-resolution grids that occur in our adaptive downsampling. 
\emph{(4)} Thanks to carefully designing the proposed method to satisfy certain guarantees, it can be used in a plug-and-play fashion within existing backbones even without retraining, making it exceptionally versatile and practical.
\emph{(5)} We further empirically show with two computer vision tasks that our method is an effective and application-agnostic tool to improve the \tradeoff\ of different established CNNs that build on regular downsampling.

%% file: sections/related_work.tex
\myparagraphnospace{Adaptive (down)sampling.}
The goal of adaptive sampling approaches is to sample an image or feature map spatially to obtain optimal performance or properties of a CNN. 
An early instance are Spatial Transformer Networks~\cite{Jaderberg:2015:STN}, which aim to increase invariance to different geometric transformations by learning to spatially transform feature maps. 
Deformable convolutions~\cite{Dai:2017:DCN} use two-dimensional offsets to the regular sampling locations of standard convolutions to allow for more flexible handling of geometric variations and different receptive field sizes within one layer.
CF-ViT~\cite{Chen:2023:CFV} entails a multi-stage approach where in each stage the input patch resolution of the most important patches is increased to refine the result of a vision transformer.
Talebi and Milanfar~\cite{Talebi:2021:LRI} show that  resizing images with learned resizers, instead of linear ones, can improve the accuracy of recognition models.
Recasens \etal~\cite{Recasens:2018:LZS} use an auxiliary saliency network to estimate the most important image regions that are then used for non-uniformly downsampling the input image, leading to an amplification of salient regions. 
Marin \etal~\cite{Marin:2019:ESL} train an auxiliary network to predict a non-uniform sampling grid that is denser near semantic boundaries and use it to non-uniformly downsample input images to retain fine detail for segmentation tasks.
Jin \etal~\cite{Jin:2022:LDS} propose a similar idea as \cite{Marin:2019:ESL}, particularly for ultra-high-resolution images, additionally including the segmentation accuracy in the training objective of the sampling grid estimator. 
Our proposed approach is fundamentally different from the above methods that adaptively downsample the image \emph{before} passing it through a CNN~\cite{Marin:2019:ESL, Jin:2022:LDS, Recasens:2018:LZS}, instead of \emph{within} the CNN. As a result, they severely sacrifice accuracy, making them primarily appropriate for very high-resolution images.

\myparagraph{Detail-preserving pooling.} While above adaptive downsampling methods are concerned with the question of \emph{how many} features to sample per region, detail-preserving pooling is concerned with the question of \emph{which} features or feature combinations to sample to better retain important detail.
Mixed Pooling~\cite{Yu:2014:MPC} randomly selects max or average pooling. 
Lee \etal~\cite{Lee:2016:GPF} learn a weighted combination of max and average pooling that is dependent on the pooling region. 
In $L_p$ pooling~\cite{Gulcehre:2014:LNP}, orders $p_j$ are learned to interpolate between different pooling operators with $p_j = 1$ corresponding to average pooling and $p_j = \infty$ to max pooling. 
Detail-preserving pooling~\cite{Saeedan:2018:DPP} is a learnable adaptive pooling method that magnifies important detail, making use of inverse bilateral filters. 
Local importance-based pooling~\cite{Gao:2019:LIP} preserves discriminative detail by training an auxiliary network to predict adaptive importance maps that are used to aggregate features for downsampling. 
 SoftPool~\cite{Stergiou:2021:RAD} minimizes information loss using a softmax-weighted sum of feature activations.
Note that adaptive downsampling methods~\cite{Jin:2022:LDS, Recasens:2018:LZS}, including our approach, also incorporate regular pooling layers, and thus, advanced pooling operations such as the above can be used complementarily.

\myparagraph{Layer aggregation and architecture.}
Besides the above, one can also use early high-resolution feature maps of CNNs and more complex network architectures to improve the granularity of feature maps or the output. Hypercolumns~\cite{Hariharan:2015:HOS} build a feature vector for any pixel by concatenating activations from all feature maps above that pixel. 
Fully convolutional networks~\cite{Long:2015:FCN} refine semantic segmentations by combining earlier layers of higher resolution with later layers of lower resolution.
The U-Net~\cite{Ronneberger:2015:UNC} architecture extends the previous idea by introducing skip connections between all layers of corresponding resolutions. 
Pinheiro \etal~\cite{Pinheiro:2016:LRO} propose to gradually refine the output segmentation by adding information from earlier layers in a top-down fashion. 
Feature pyramid networks~\cite{Lin:2017:FPN} produce feature pyramids by combining high-level and low-level features of a CNN.
Instead of simple one-step skip-connections, Yu \etal~\cite{Yu:2018:DLA} incorporate more depth and sharing in their layer aggregation.
HRNet~\cite{Wang:2015:DHR} goes even further and processes high-resolution and low-resolution streams in parallel while repeatedly exchanging information between the streams.
Yu \etal~\cite{Yu:2017:DRN} improve the granularity of feature maps by substituting downsampling operations with dilated convolutions~\cite{Yu:2016:MSC}, allowing to retain the feature map resolution while keeping the original receptive field of the CNN.
Generally, high-level feature maps yield semantically stronger features~\cite{He:2016:DRL, Lin:2017:FPN}, and thus, using auxiliary information from low-level layers might not be optimal. 
Further, combining layers of different resolutions or keeping higher resolutions can potentially introduce new parameters, increase computational cost, and/or raise memory usage.

\myparagraph{Sparse convolution.}
While sparse convolution is not directly related to adaptive downsampling, it plays an essential role in our method.
Initial works on sparse convolution~\cite{Engelcke:2017:V3D, Graham:2014:SSC, Graham:2015:S3D, Riegler:2017:ONL} improve the computational cost of standard convolution by processing only active elements in sparse inputs, such as point clouds~\cite{Engelcke:2017:V3D}, handwritten digits~\cite{Graham:2014:SSC}, or 3D grids~\cite{Riegler:2017:ONL}.
Submanifold sparse convolution~\cite{Graham:2018:3DS} avoids a growing number of active elements caused by standard convolutions that dilate the sparse data with each layer. 
Contrary to sparse convolution, which is only used to completely ignore inactive elements, we adapt sparse convolution to work with our proposed multi-resolution feature maps.

%% file: sections/method.tex
\newcommand{\downsamplingfactor}{d}
\newcommand{\featuremapdown}{f_d}

In this work, we postulate that not all feature map regions of a CNN are equally important and that, therefore, different regions should be processed at \emph{different resolutions}. By processing only a subset of the most important feature map regions at higher resolution while downsampling less important ones, we can combine the smaller representation size of regular downsampling with the higher feature map granularity of dilated convolutions~\cite{Yu:2017:DRN}.
As a consequence, we can gain relatively large improvements in predictive performance while only moderately increasing the computational cost compared to regular downsampling.

The high-level idea behind our approach is illustrated in \cref{fig:fig1_right}(b). Contrary to regular downsampling (\cref{fig:fig1_right}(a)), where the image or feature map is downsampled uniformly by sampling every second pixel, we utilize an adaptive downsampling scheme (\cref{fig:fig1_right}(b)) that retains higher resolution in areas with fine detail while downsampling less informative ones. 
In the following, we will outline the details of our method with a focus on two-dimensional feature maps with channels, \eg, in CNNs for vision tasks.
However, the extension to other dimensionalities, such as 1D signals and 3D temporal or volumetric data, works analogously.

\myparagraph{Regular downsampling.}
Let $f \in \R^{H \times W \times C}$ be a feature map (or image) of height $H$, width $W$, and depth $C$. 
We denote regular downsampling by a factor of $\downsamplingfactor \in \mathbb{N}_{>1}$ as the process of reducing the spatial dimension of $f$ by outputting one element $e \in \R^C$ from each non-overlapping $d \times d$ patch, such that the resulting feature map $\featuremapdown$ is of shape $\frac{H}{d} \times \frac{W}{d} \times C$. 
Generally, regular downsampling ${\pool \colon \R^{H \times W \times C} \mapsto \R^{H/d \times W/d \times C}}$ in CNNs is realized by applying a patch-wise downsampling function $\pool^\prime \colon \R^{d \times d \times C} \mapsto \R^C$ to all $HW/d^2$ non-overlapping $\downsamplingfactor \times \downsamplingfactor$ patches $f^\prime_{i,j}$ in $f$:
\begin{equation}
\featuremapdown = \pool \left( f \right) = \big(\pool^\prime (f^\prime_{1,1}), \ldots , \pool^\prime (f^\prime_{H/d,W/d})\big) \, .
\end{equation}
The above definition of regular downsampling generalizes to many commonly used downsampling methods used in CNNs. 
For example, $\pool^\prime$ can take the form of various pooling operations, \eg,~\cite{Gulcehre:2014:LNP, Saeedan:2018:DPP, Yu:2014:MPC}, or it can select an element based on its spatial location to implement uniform downsampling, \ie, sampling every $d$-th element. When following a convolution of stride 1, this corresponds conceptually to a strided convolution with a stride of $d$.

\myparagraph{Adaptive downsampling.}
In this work, we generalize the above regular downsampling scheme by allowing some patches $f^\prime_{i,j}$ to \emph{not be downsampled}, and therefore, retaining their \emph{original resolution} of $d\times d$.
This is accomplished by providing an additional downsampling mask $m \in \{0,1\}^{H/d \times W/d}$ as input, specifying which patches $f^\prime_{i,j}$ should be downsampled ($m_{i,j}=1$) and which not ($m_{i,j}=0$). 
For now, we assume that this downsampling mask is given, and go later into more detail about how to compute it in a content-adaptive way.
Formally, our novel adaptive downsampling $\pooladaptive$ is defined as
\begin{equation}
\begin{split}
\pooladaptive \left( f,m \right) &= (\pooladaptive^\prime \big(f^\prime_{1,1}), \ldots , \pooladaptive^\prime (f^\prime_{H/d,W/d})\big)\, , \\
&\text{with}\ 
\pooladaptive^\prime (f^\prime_{i,j}) = 
\begin{cases}
   f^\prime_{i,j} & \text{if}\ m_{i,j} = 0\\
   \pool^\prime (f_{i,j}^\prime) & \text{if}\ m_{i,j} = 1 \, ,
\end{cases}
\end{split}
\label{eq:adaptive_downsample}
\end{equation}
denoting a patch-wise adaptive downsampling as indicated by the downsampling mask $m$.
\begin{figure*}
  \centering
  \includegraphics[width=0.975\linewidth]{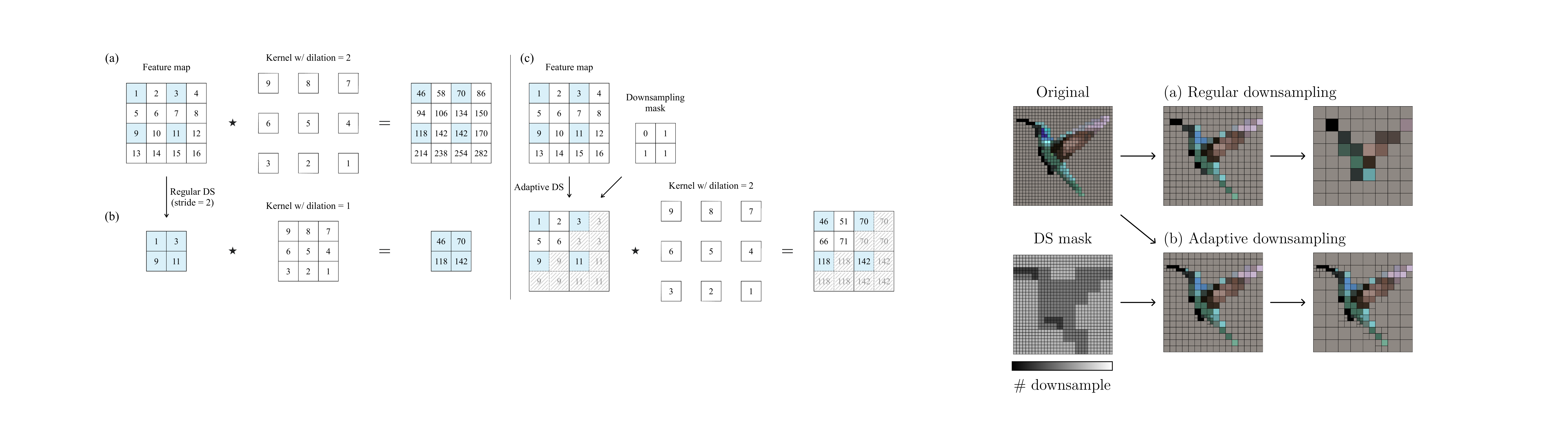}
  \vspace{-5pt}
  \caption[Figure 2]{\emph{Illustration of different methods to handle feature map resolution in CNNs.} $\star$ denotes the two-dimensional cross-correlation operator (w/ zero padding). \emph{(a)} Dilated convolution increases the receptive field (equally to downsampling) while retaining the feature resolution \cite{Yu:2017:DRN}. \emph{(b)} Standard strided convolution in CNNs (with stride=2) corresponds to uniformly downsampling the feature map by sampling every second element. This is typically followed by a standard convolution. \emph{(c)} In our adaptive scheme, the downsampling mask is used to downsample a subset of the feature map and low-resolution entries are projected into the high-resolution grid; shaded cells (hatched) denote inactive elements. Afterward, sparse convolution is applied to the active elements (black font) of the multi-resolution feature map to produce an output with the same, multiple resolutions as the input. Note how the (loop) invariant of all algorithms (Guarantee~2) ensures that the elements highlighted in blue, both before and after convolution, are the same across the different methods.} \label{fig:adaptive_pool_illu}
  \vspace{-0.25em}
\end{figure*}

\myparagraph{Multi-resolution grid convolution.}
As the resulting feature map $\pooladaptive \left( f,m \right)$ consists of multiple resolutions, we lose the regular grid structure of our data necessary to work with standard convolutional layers.
To address this, we project the lower-resolution features onto their corresponding location in the high-resolution grid, as illustrated in \cref{fig:adaptive_pool_illu}(c). Naturally, high-resolution patches $f^\prime_{i,j}$ that are filled with a lower-resolution element will have empty cells (\cref{fig:adaptive_pool_illu}(c) -- shaded cells) and, thus, are sparse. 
We exploit the resulting sparsity by processing our feature representations in subsequent layers using sparse convolution~\cite{Engelcke:2017:V3D, Graham:2015:S3D, Riegler:2017:ONL}.
This allows to only consider \emph{active} elements, and hence, uses less computation in regions of lower effective resolution.
To avoid submanifold dilation~\cite{Graham:2018:3DS}, \ie, that inactive elements become active after convolution, we suggest a modified instance of submanifold sparse convolution~\cite{Graham:2018:3DS} (SSC). SSC ensures that input and output contain the same set of active elements and thus enables us to keep the resolution of each respective feature map or image region unchanged.
To retain the same receptive field as the convolutional layer would have with regular downsampling, we increase the dilation factor to $d$.
A remaining challenge is that at specific feature map locations (see, \eg, \cref{fig:adaptive_pool_illu}(c) -- cell with $6$) non-central elements of the convolutional kernel could fall onto inactive elements, causing the result to be corrupted. 
To avoid this, we modify SSC to assume that all inactive elements take the value of the corresponding active element within their patch (shown in \cref{fig:adaptive_pool_illu}(c) as shaded cells). 

\myparagraph{Extension to CNNs.}
Similarly to Yu \etal~\cite{Yu:2017:DRN}, our proposed adaptive downsampling is incorporated into a standard CNN with regular downsampling, \ie, pooling or strided convolution, by substituting the last $n$ downsampling operations with adaptive downsampling.
Further, we substitute all convolutional layers that follow the $i \in \{1,\ldots,n\}$ adaptive downsampling operations with our proposed sparse convolution with a dilation factor of $d^i$ to keep the same receptive field as the original network. 
To perform multiple adaptive downsampling operations in succession, we only consider the elements belonging to the currently lowest available resolution in our multi-resolution feature map and convert them into a dense representation. We then perform adaptive downsampling on these elements according to \cref{eq:adaptive_downsample}. As before, the resulting features are projected to their corresponding location in the multi-resolution feature map.
Consequently, patches already containing higher-resolution elements are excluded from the downsampling process.
This implies that areas, where resolution is retained, cannot be downsampled at a later downsampling step, and thus, for $d=2$, one obtains a quadtree-like resolution pattern as seen in \cref{fig:fig1_right}(b).
A more detailed explanation of this process can be found in the supplementary.

\begin{table*}
    \caption[Table 1]{\emph{Oracle experiment for semantic segmentation.} For each model and output stride (OS), we report the mean IoU (mIoU) on the Cityscapes~\cite{Cordts:2016:CDS} evaluation split, the theoretical (average) number of multiply-adds (\#MA), and the min./max.~inference times in seconds. Our adaptive downsampling (OS=$8$$\rightarrow$$32$) improves the \tradeoff{} of all models, indicated by the on par mIoU scores and the significantly reduced \#MA and inference times, compared to OS=8.
    }
    \vspace{-1em} 
    \label{table:exp1}
    \smallskip
    \centering
    \footnotesize
    \begin{tabular*}{\textwidth}{@{\extracolsep{\fill}} lccccccccc}
        \toprule
        \multicolumn{1}{c}{} & \multicolumn{9}{c}{\textbf{Different backbones with DeepLabv3~\cite{Chen:2017:RAC}}}
        \\
        \cmidrule(lr){2-10}
        \multicolumn{1}{c}{} & \multicolumn{3}{c}{ResNet-50~\cite{He:2016:DRL}} & \multicolumn{3}{c}{ResNet-101~\cite{He:2016:DRL}} & \multicolumn{3}{c}{ResNet-152~\cite{He:2016:DRL}} \\
        \cmidrule(lr){2-4}\cmidrule(lr){5-7}\cmidrule(lr){8-10}
        OS & $16$ & $8$ & $8$$\rightarrow$$32$ & $16$ & $8$ & $8$$\rightarrow$$32$ & $16$ & $8$ & $8$$\rightarrow$$32$\\
       \midrule
        mIoU & $0.7496$ & $0.7655$ & $0.7658$ & $0.7591$ & $0.7775$ & $0.7748$ & $0.7683$ & $0.7852$ & $0.7848$ \\
        \#MA & $2.6e11$ & $8.0e11$ & $3.8e11$ & $4.2e11$ & $1.4e12$ & $6.7e11$ & $5.7e11$ & $1.9e12$ & $1.0e12$ \\
        Time & $0.05$ & $0.11$ & $0.05/0.08$ & $0.07$ & $0.22$ & $0.08/0.15$ & $0.1$ & $0.31$ & $0.11/0.25$\\
\end{tabular*}
\vspace{-1.2em}    
\end{table*}        
       
\begin{table*}
    \centering
    \footnotesize
    \begin{tabular*}{\textwidth}{@{\extracolsep{\fill}} lcccccccccccc}
        \toprule        \multicolumn{1}{c}{} & \multicolumn{6}{c}{\textbf{ResNet-101~\cite{He:2016:DRL} with different segmentation heads}} & \multicolumn{6}{c}{\textbf{ResNet-101~\cite{He:2016:DRL} with DeepLabv3~\cite{Chen:2017:RAC} and different extensions}}
        \\
        \cmidrule(lr){2-7}\cmidrule(lr){8-13}
        \multicolumn{1}{c}{} & \multicolumn{3}{c}{FCN~\cite{Long:2015:FCN}} & \multicolumn{3}{c}{DeepLabv3+~\cite{Chen:2018:ECA}} & \multicolumn{3}{c}{SoftPool~\cite{Stergiou:2021:RAD}} & \multicolumn{3}{c}{Deformable Convolution~\cite{Dai:2017:DCN}}\\
        \cmidrule(lr){2-4}\cmidrule(lr){5-7}\cmidrule(lr){8-10}\cmidrule(lr){11-13}
        OS & $16$ & $8$ & $8$$\rightarrow$$32$ & $16$ & $8$ & $8$$\rightarrow$$32$ & $16$ & $8$ & $8$$\rightarrow$$32$ & $16$ & $8$ & $8$$\rightarrow$$32$\\
        
        \midrule
        mIoU & $0.7154$ & $0.7292$ & $0.7322$ & $0.7745$ & $0.781$ & $0.7805$ & $0.7652$ & $0.7823$ & $0.7819$ & $0.7564$ & $0.7801$ & $0.7797$\\
        \#MA & $4.2e11$ & $1.4e12$ & $6.7e11$ & $4.2e11$ & $1.4e12$ & $6.7e11$ & $4.2e11$ & $1.4e12$ & $7.4e11$ & $8.0e11$ & $1.4e12$ & $1.0e12$\\
        Time & $0.07$ & $0.22$ & $0.08/0.15$ & $0.07$ & $0.22$ & $0.08/0.15$ & $0.07$ & $0.22$ & $0.08/0.17$ & $0.1$ & $0.21$ & $0.11/0.17$\\
        
        \bottomrule
    \end{tabular*}
    \vspace{-0.25em}
\end{table*}

\myparagraph{Guarantees.}
The resulting CNN can further be seen as a generalization of both regular CNNs with downsampling as well as of the dilated convolution of Yu \etal~\cite{Yu:2017:DRN}.
When setting all elements of the downsampling mask to one, we obtain a standard CNN with regular downsampling.
When setting all elements of the downsampling mask to zero, we obtain exactly dilated convolution \cite{Yu:2017:DRN} for higher-resolution feature maps. 
By simultaneously processing different feature map regions at \emph{different} resolutions, controlled by the downsampling mask, we establish a combination of the above methods that allows us to ``interpolate'' between different output strides in a \emph{locally adaptive} fashion. 
An important desideratum behind our approach is that -- like dilated convolution \cite{Yu:2017:DRN}  -- additional training of the backbone is not mandatory.
Therefore, it is exceptionally easy to use in a plug-and-play fashion. 
This is possible due to the following two guarantees, which result from our design and hold when substituting regular downsampling, \ie, strided convolution or pooling, with our adaptive downsampling:
\begin{itemize}
  \item[] \textbf{Guarantee 1:} Due to projecting all elements into the highest-resolution grid and adjusting the dilation factors of subsequent sparse convolutions, we can guarantee that the receptive fields of the CNN and each intermediate layer remain unchanged compared to the corresponding CNN with regular downsampling.
  \item[] \textbf{Guarantee 2:} All output features of a standard CNN with strided convolution equal the output features at the corresponding high-resolution locations of the same CNN with adaptive downsampling (see \cref{fig:adaptive_pool_illu}(c) -- blue cells).
\end{itemize}
This means that a CNN with adaptive downsampling is not modifying feature map values of a standard CNN with strided convolution, but instead only \emph{adds} information at pixel locations where there was none before, and therefore, refines the feature map. 
Note that while Guarantee 2 formally only holds for strided convolutions and not for pooling, we show in \cref{exp:case_keypoint} that applying our adaptive downsampling in a CNN with pooling is still feasible in practice without necessarily requiring retraining.

\myparagraph{Content-adaptive downsampling masks.}
A remaining question is how to obtain the downsampling mask to indicate what regions to process at which resolution in a locally content-adaptive way. 
In \cref{sec:experiments}, we show that traditional algorithms, like high-frequency detection or keypoint estimates, can serve as an effective basis to identify important image regions, and thus, to estimate downsampling masks.

Additionally, we demonstrate how to learn an appropriate downsampling mask from data. Specifically, we utilize a shallow CNN that takes as input the feature map of the main model before the adaptive downsampling step, and outputs the downsampling mask. In order to train it end-to-end with the main model, we make the discrete mask estimation differentiable with Gumbel-Softmax~\cite{Jang:2017:CRG}. As training our adaptive downsampling end-to-end can result in a downsampling mask of only zeros, \ie, full resolution, we control the amount of invested resources with an additional hyperparameter $\gamma \in [0,1]$ that defines the desired proportion of active mask elements. Its squared difference to the actual proportion $\hat{m} \in [0,1]$ of active elements in our downsampling mask is included in the final loss function
\begin{equation}
    \mathcal{L} = \alpha\mathcal{L}_1 + \beta(\gamma - \hat{m})^2,
\end{equation}
with $\alpha$ and $\beta$ being weighting factors, and $\mathcal{L}_1$ denoting the loss of the main task, \eg, the segmentation loss.

\myparagraph{Limitations.}
Our approach allows to interpolate between different feature map resolutions, which can drastically improve the \tradeoff\ as shown in \cref{sec:experiments}.
However, some limitations arise that could hinder its effective usage in some applications. 
First, our method's benefit depends on the given downsampling mask.
If the mask does not capture important regions, retaining higher-resolution regions will not help. 
Second, there may be backbones, especially those with advanced layer aggregation schemes~\cite{Wang:2015:DHR, Yu:2018:DLA}, or datasets without fine details where higher-resolution feature maps and, therefore, also our method generally do not yield advantages. 
However, our analysis and related work~\cite{Yu:2017:DRN, Dusmanu:2019:D2N} show that the highly impactful and widely used CNN backbones, VGG~\cite{Simonyan:2015:VDC} and ResNet~\cite{He:2016:DRL}, benefit from higher-resolution feature representations in various applications.

%% file: sections/experiments.tex
To demonstrate the practicality of our proposed adaptive downsampling scheme, we conduct various experiments on two different computer vision tasks confirming the following points: \emph{(1)} As a generalization of \cite{Yu:2017:DRN} and standard CNNs, our method allows to interpolate between different output strides, enabling a more granular control of how many resources should be allocated for the task and input image at hand. \emph{(2)} By selecting task-relevant regions for a specific input, our method can drastically improve the \tradeoff\ of standard CNNs.
\emph{(3)} The guarantees satisfied by our method allow to incorporate it in a plug-and-play fashion into pre-trained networks at test time. 
\emph{(4)} Our method is not limited to a single application and finding appropriate downsampling masks is feasible in practice.
\emph{(5)} Our approach generalizes to different backbones, segmentation heads, and extensions. 

We start our evaluation with an oracle experiment, demonstrating the potential of our method given a high-quality downsampling mask. Afterward, we present two case studies that show how our method can be used in practice, demonstrating the advantages summarized above.

\myparagraph{Experimental setup (segmentation).} 
Experiments for semantic segmentation (\cref{exp:oracle_seg,exp:case_seg}) have been conducted on the Cityscapes dataset~\cite{Cordts:2016:CDS}. 
We report the mean intersection over union (mIoU) on the validation set.

\myparagraph{Experimental setup (keypoint description).} 
For the keypoint description experiment (\cref{exp:case_keypoint}), we use the established D2-Net descriptor~\cite{Dusmanu:2019:D2N}, which takes an image as input and outputs a dense feature map that can be used to locate and describe keypoints.
Since D2-Net keypoint localization tends to be imprecise~\cite{Uzpak:2020:STK}, we follow Uzpak \etal~\cite{Uzpak:2020:STK} and instead use SIFT~\cite{Lowe:2004:DIF} to first detect the 512 most salient keypoints and employ D2-Net only to describe them.
Following common practice~\cite{Luo:2020:ALL, Revaud:2019:R2D}, we report the mean matching accuracy, \ie, the ratio between correct and possible matches, at a three-pixel threshold (MMA@3) on the HPatches dataset~\cite{Balntas:2017:HBE}.

\myparagraph{Experimental setup (general).} As the focus of this work is the \tradeoff{} of the examined methods, for all experiments we report the respective task-specific metrics, \ie, mIoU and MMA@3, over the required theoretical (average) number of multiply-adds as a proxy of computational cost. 
As baselines, we consider the original models with varying output strides obtained by using regular downsampling~\cite{Simonyan:2015:VDC, He:2016:DRL} with the standard factor $d=2$, respectively \emph{not} downsampling and using dilated convolutions~\cite{Yu:2017:DRN}.
To draw a more conclusive picture, we also report the required inference time in seconds. However, the inference time depends on the used implementation and hardware -- here an Nvidia RTX A6000 GPU. For a fair comparison we use the same implementation for our method and the baselines; please refer to the supplementary for different implementations.

\begin{figure}[t!]
  \centering
  \includegraphics[width=1.\linewidth]{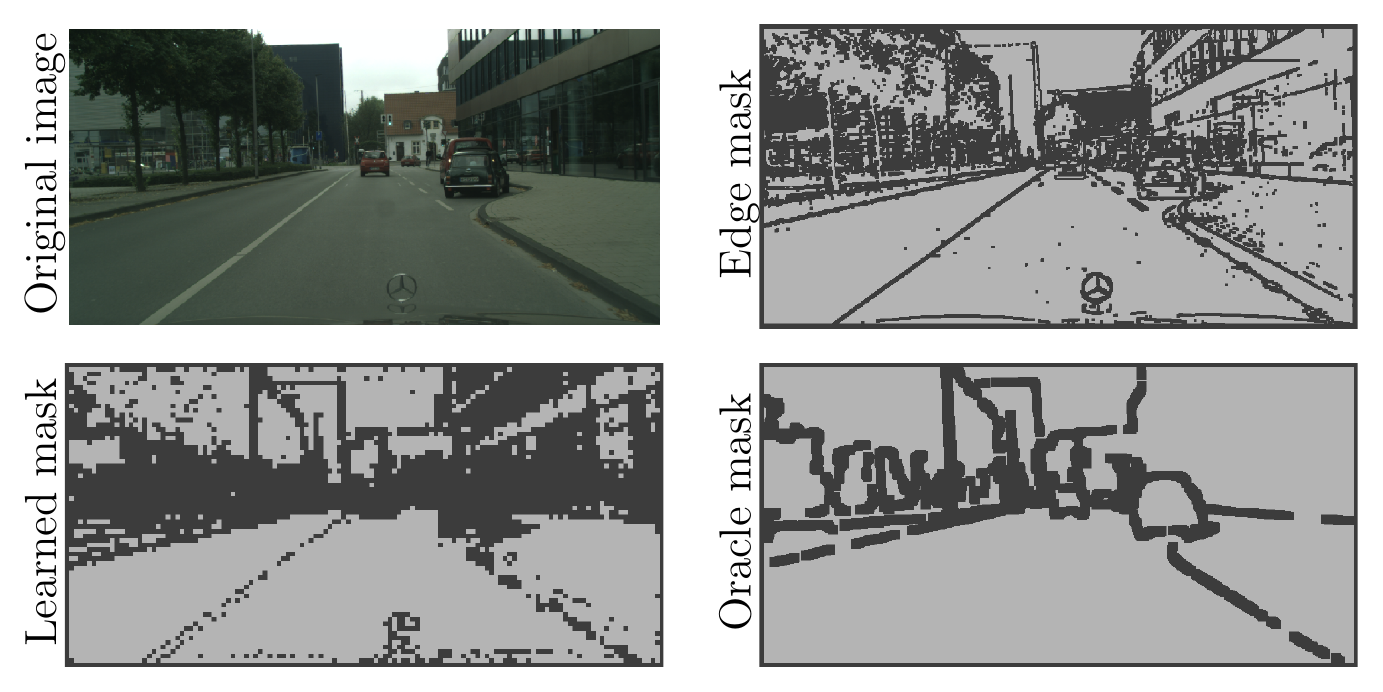}
  \vspace{-20pt}
  \caption[Figure 5]{
  \emph{Example downsampling masks.} A Cityscapes~\cite{Cordts:2016:CDS} image and the corresponding estimated downsampling masks using our oracle setup, edge detection, and a shallow learned network. Note how low-frequency regions, \eg, the sky and road, belong to the same class, and thus, can be processed at low resolution (bright) to decrease the computational cost.}
  \label{fig:seg2_masks}
  \vspace{-0.25em}
\end{figure}

\subsection{Oracle experiment: Feature resolution in semantic segmentation}\label{exp:oracle_seg}

In our first experiment, we examine the role of feature map resolution in semantic segmentation and whether processing more informative regions at a higher resolution offers advantages and is generally possible. To this end, we first conduct an oracle experiment in which we assume a high-quality downsampling mask to be given. 
As baselines we use a segmentation model with regular downsampling and two different output strides (OS) of 16 (lower feature resolution) and 8 (higher feature resolution). For our proposed adaptive downsampling with output stride $8$ to $32$ (OS=$8$$\rightarrow$$32$), we define the downsampling mask as the pixels that are misclassified by a regular model with OS=$32$ but correctly classified by a model with OS=$8$, followed by a dilation with a square kernel size of $K\times K$ (see \cref{fig:seg2_masks} and supplementary). This gives us exactly the image regions that benefit from higher resolutions, and thus, can be considered our oracle downsampling mask. To demonstrate the generalizability of our approach, we investigate a variety of different backbones, segmentation heads, and extensions.

Results for our adaptive downsampling and baselines can be seen in \cref{table:exp1}.
Confirming our assumption and prior work~\cite{Yu:2017:DRN, Chen:2018:ECA}, a smaller output stride, respectively higher resolution, increases both the mIoU and computational cost of all the examined baseline models (OS=$8$ \vs OS=$16$).
Moreover, using our adaptive downsampling leads to an mIoU that is on par with the respective regular model with OS=$8$ while for some images requiring less than 50\% of the time and computational cost (\eg, ResNet-101+DeepLabv3). This clearly confirms our hypothesis that only a fraction of the feature map must be processed at a higher resolution to obtain strong predictive performance. Further, it shows that our proposed method is applicable to various different ResNet backbones (VGG~\cite{Simonyan:2015:VDC} is shown in \cref{exp:case_keypoint}) and segmentation heads. Also, our approach is a novel and orthogonal research direction, which can be used \textit{together} with existing extensions like deformable convolution~\cite{Dai:2017:DCN} and advanced pooling strategies~\cite{Stergiou:2021:RAD}.

\begin{figure}
  \centering
  \includegraphics[width=0.9\linewidth]{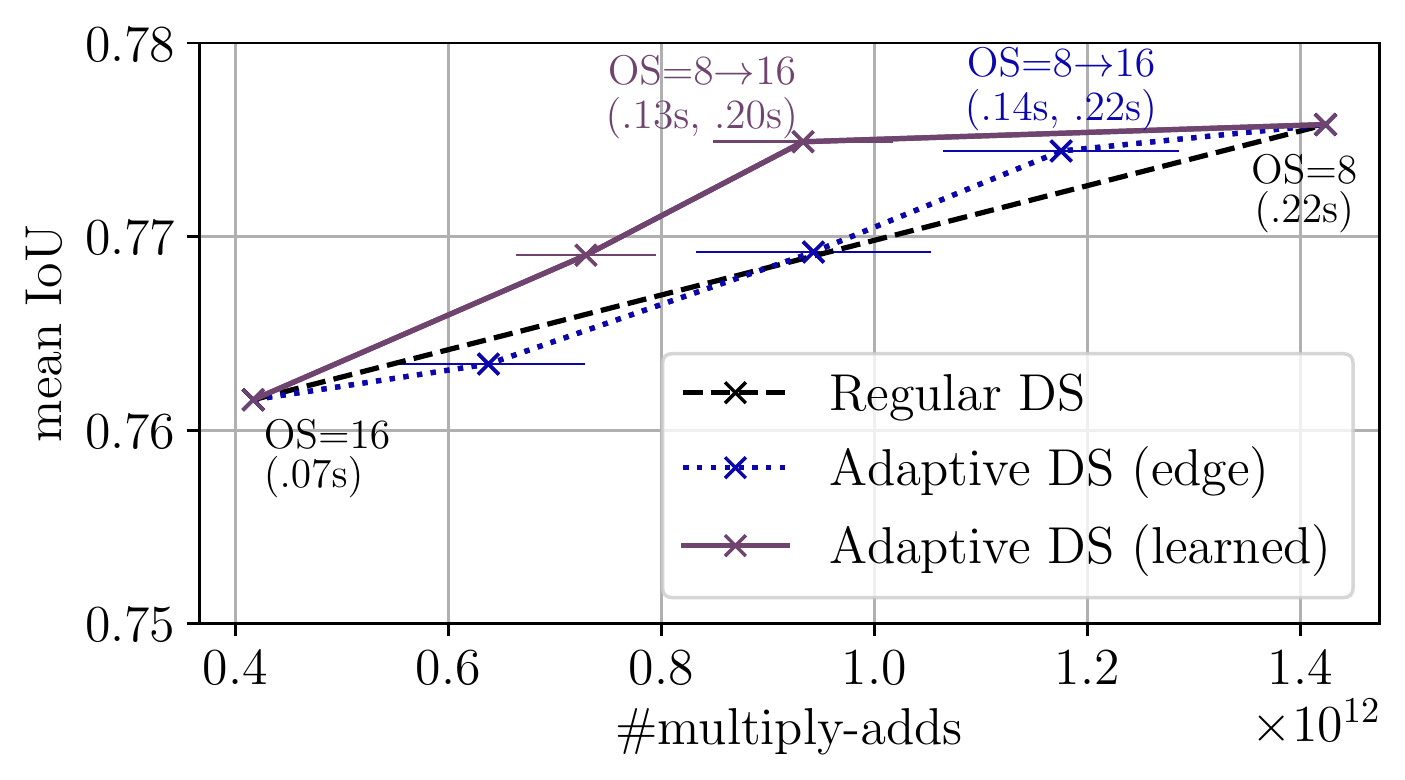}
  \vspace{-6pt}
  \caption[Figure 4]{\emph{Semantic segmentation results.}
  mIoU on the Cityscapes~\cite{Cordts:2016:CDS} evaluation set over the number of multiply-adds for a ResNet-101~\cite{He:2016:DRL} backbone with a DeepLabv3~\cite{Chen:2017:RAC} segmentation head, using regular downsampling~\cite{Yu:2017:DRN} and our novel adaptive downsampling scheme with a learned mask, respectively an edge detection mask, with varying hyperparameters. Min.~and max.~inference times for each setup are annotated.
  To reduce variance, we report the maximum mIoU over three runs. X-confidence intervals show 2 times the standard deviation of the required multiply-adds. 
  }
  \label{fig:seg2}
  \vspace{-0.25em}
\end{figure}

\subsection{Case study 1: Semantic segmentation}\label{exp:case_seg}

In our first case study, we again consider semantic segmentation with ResNet101+DeepLabv3 models but now with realistically estimated downsampling masks.
We propose two strategies to estimate useful downsampling masks:

\myparagraph{Edge mask.} From analyzing the problem at hand, we know that segmentation boundaries are likely to align with edges.
For this reason, we perform edge detection on the input image to create a downsampling mask.
Specifically, we apply a Sobel filter to the original grayscale images and dilate the thresholded result with a square kernel of size $11 \times 11$. Computational cost for detecting the edges is neglectable ($3\times10^{-3}\%$ of the total multiply-adds).

\myparagraph{Learned mask.} As described in \cref{sec:methods}, we use a shallow CNN to estimate downsampling masks from an intermediate layer of the feature extractor. The mask estimator is trained end-to-end with the segmentation model (see supplementary). The computational cost for the mask estimator is included in the reported multiply-adds and times.

\begin{figure}[t!]
  \centering
  \includegraphics[width=1.0\linewidth]{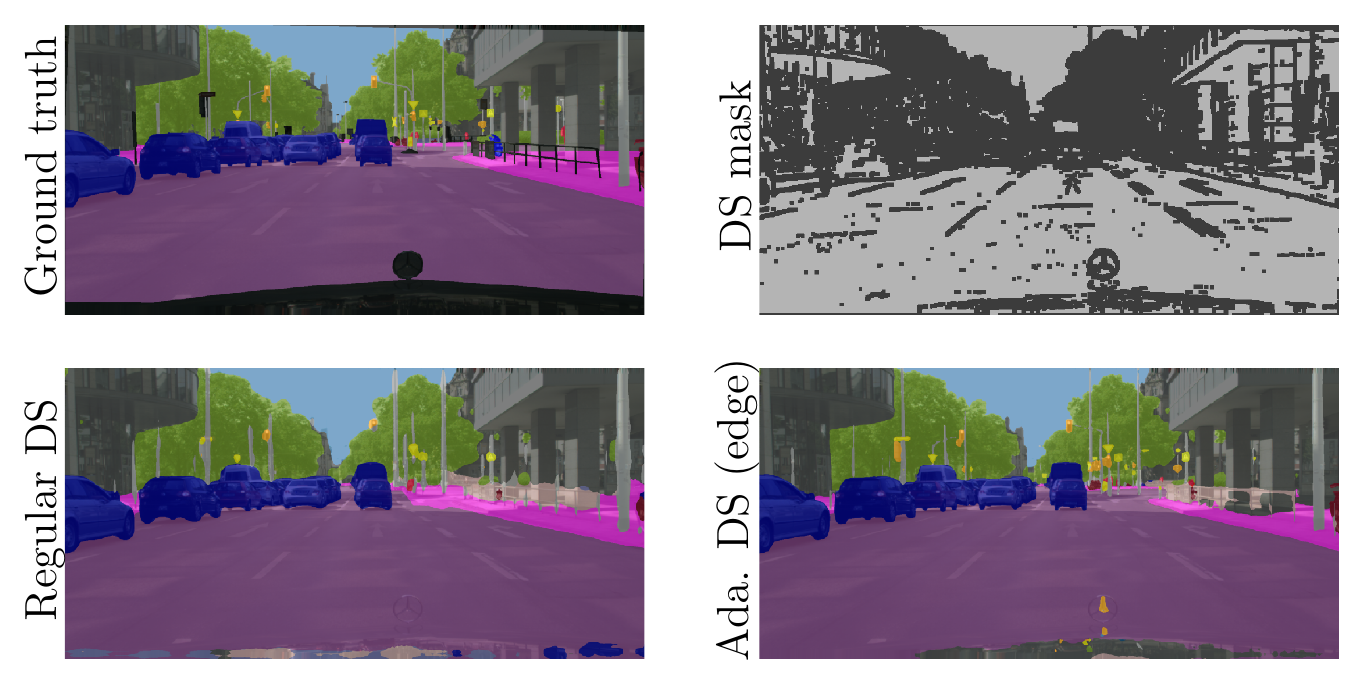}
  \vspace{-20pt}
  \caption[Figure 5]{
  \emph{Qualitative semantic segmentation results.} Estimated segmentation maps using regular downsampling with OS=$8$ and adaptive downsampling with an output stride of 8 to 16 (OS=$8$$\rightarrow$$16$) and our edge downsampling (DS) mask.}
  \label{fig:seg2_results}
  \vspace{-0.25em}
\end{figure}

\smallskip
The resulting downsampling masks for a single example image can be seen in \cref{fig:seg2_masks}. 
Dark areas denote keeping the higher resolution of an output stride of 8 while bright areas correspond to an output stride of 16 (OS=$8$$\rightarrow$$16$).
In \cref{fig:seg2}, we report results for our adaptive downsampling with the two proposed strategies.
We again observe that by ``focusing'' on more relevant image regions, \eg, edges, we can improve the \tradeoff{} of an established model. Further, by adjusting the hyperparameters of the mask estimators, we can granularly control the amount of invested resources.
Remarkably, adaptive downsampling with a learned mask achieves on par mIoU as regular downsampling with an output stride of 8 (0.775~vs.~0.776) while reducing the required computational cost by approximately 35\%. Looking at the annotated minimum and maximum times, we can observe time improvements of up to $40\%$. However, we observe that for images with a lot of ``interesting'' content, the worst-case inference time is similar to regular downsampling with OS=8.
This case study also demonstrates how naively and easily we can estimate downsampling masks that together with our adaptive downsampling still yield advantages for the task at hand.
A visual example of the estimated segmentations is shown in~\cref{fig:seg2_results}.

Contrary to our work, existing methods that adaptively downsample the input \emph{before} passing it into a CNN~\cite{Marin:2019:ESL, Jin:2022:LDS} lose important image information, resulting in lower mIoU scores of $0.5$ or $0.65$ on Cityscapes~\cite{Cordts:2016:CDS}. Hence, they are mainly beneficial for processing ultra-high resolution images and are not considered as competitive baselines here.

\subsection{Case study 2: Keypoint description}\label{exp:case_keypoint}

\begin{figure}
  \centering
  \includegraphics[width=0.9\linewidth]{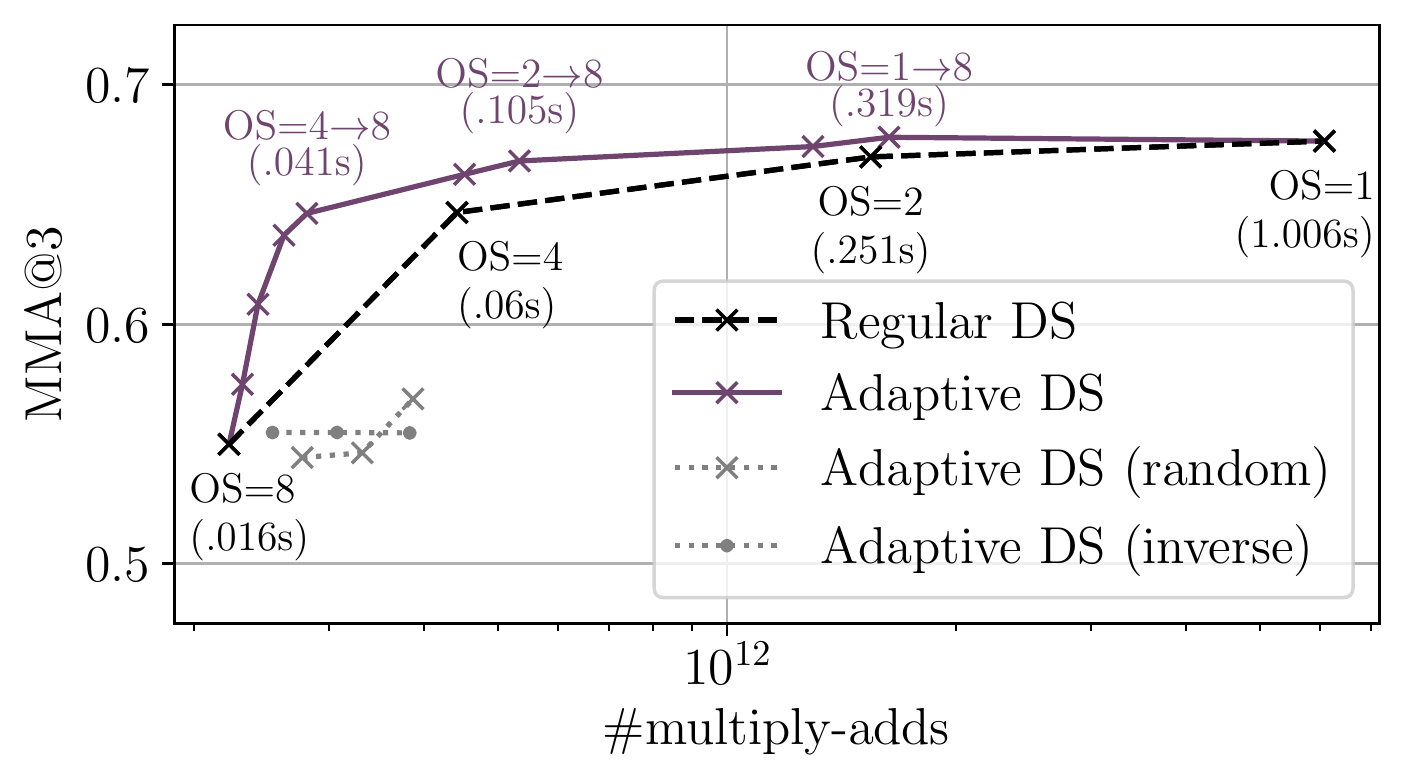}
  \vspace{-6pt}
  \caption[Figure 6]{\textit{Keypoint description results.} MMA@3 over the required number of multiply-adds for a SIFT~\cite{Lowe:2004:DIF} detector with a pre-trained D2-Net~\cite{Dusmanu:2019:D2N} descriptor using regular downsampling~\cite{Yu:2017:DRN} with different output strides (OS), and our novel adaptive downsampling scheme using different masks. Given a reasonable downsampling mask, our adaptive downsampling can achieve on par MMA@3 while significantly reducing the computational cost. 
  Actual inference times for a randomly chosen image are annotated.} \label{fig:keypoints_right}
  \vspace{-25pt}
\end{figure}

In our second realistic case study, we consider the problem of keypoint description.
To this end, we investigate the established SIFT detector~\cite{Lowe:2004:DIF} and D2-Net descriptor~\cite{Dusmanu:2019:D2N} as described in the experimental setup.
D2-Net utilizes a VGG16~\cite{Simonyan:2015:VDC} backbone that progressively applies max pooling to reduce the resolution of the feature map, and thus, computational load.
To demonstrate the simplicity and versatility of our method, we use the original model weights provided by the authors~\cite{Dusmanu:2019:D2N} and do \textit{not} perform additional training for any of the following setups. 
Again confirming prior work~\cite{Dusmanu:2019:D2N}, in \cref{fig:keypoints_right} we see that a higher feature map resolution (\ie, lower OS) increases both the MMA@3 and the computational cost of the baseline model. 

\begin{figure}
  \centering
  \includegraphics[width=0.86\linewidth]{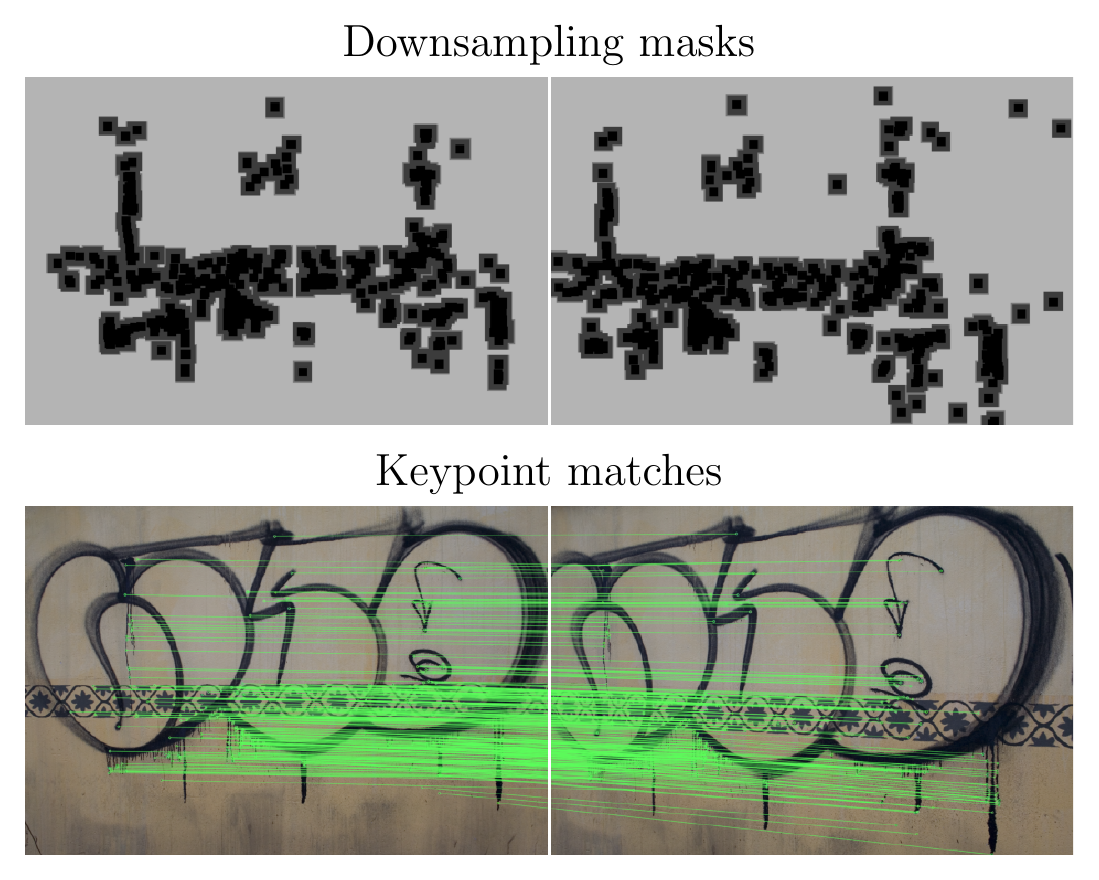}
  \vspace{-8pt}
  \caption[Figure 5]{Qualitative results of keypoints matched with our adaptive downsampling scheme and the corresponding downsampling masks. Note how only a fraction of the image is important, and thus, processed at high resolution (dark) to increase the accuracy while saving resources in bright areas of the downsampling mask.} \label{fig:keypoints_viz}
  \vspace{-0.25em}
\end{figure}

Next, we substitute the last 3, 2, or 1 max pooling operations with our adaptive downsampling with max pooling (OS=$\{1,2,4\}$$\rightarrow$$8$).
As only image regions around the keypoints are important for our task at hand, we estimate the downsampling masks by dilating keypoints obtained from SIFT with filters of increasing sizes for each of the up to three downsampling levels (see \cref{fig:keypoints_viz}).
We demonstrate a fine-grained control of computational cost by varying the dilation sizes for the downsampling levels.
Larger dilation sizes will lead to larger areas being processed at higher resolution, and thus, increase cost as well as MMA@3. 
The results in \cref{fig:keypoints_right} clearly show that, compared to the fixed output strides of the baseline model, our adaptive downsampling yields a significant reduction of the computational cost while achieving the same MMA@3, \eg, we achieve an ${\sim}70\%$ reduction of the computational cost to reach an MMA@3 that is on par with a regular output stride of 1.

\myparagraph{Inadequate masks.}
To evaluate how our method performs with ``bad'' downsampling masks, in \cref{fig:keypoints_right} we additionally report the MMA@3 for adaptive downsampling with an output stride of $4$ to $8$ using random and unreasonable masks, \ie, the inverse of reasonable masks. 
Confirming our stated limitations, we see that inadequate downsampling masks do not capture important image regions, and thus, we process unimportant regions at high resolution, leading to an increased computational cost compared to regular downsampling with OS=$8$ without yielding significant improvements in predictive performance. 
Note, however, that thanks to our guarantees in \cref{sec:methods}, poor downsampling masks still yield comparable predictive performance as regular downsampling with OS=$8$, showing that these masks do not negatively affect predictive performance and that the model still behaves in an expected way.

%% file: sections/conclusion.tex
In this work, we propose --~to the best of our knowledge~-- the first downsampling scheme that changes the operating resolution \emph{within} CNNs in a \emph{locally adaptive} fashion.
We do so by generalizing standard CNNs~\cite{Krizhevsky:2012:INC, Simonyan:2015:VDC, He:2016:DRL} and Yu \etal's~\cite{Yu:2017:DRN} dilated convolutional networks, allowing us to process feature maps with spatially-varying resolutions.
By selecting an appropriate content-adaptive downsampling mask, indicating locally the most important regions that are to be processed at a higher resolution, we can enable CNNs to ``focus'' more strongly on task-relevant regions.
Besides substantially improving the \tradeoff{} in two computer vision tasks, our novel adaptive downsampling enables a more continuous control of the invested computational resources, giving practitioners another degree of freedom to best adapt their model to the available resources.
Thanks to carefully designing the proposed method, it satisfies two important guarantees, allowing adaptive downsampling even to be used at test time in a plug-and-play fashion within standard CNNs pre-trained with \textit{regular} downsampling.
As our approach improves the \tradeoff\ of various established models, it contributes to saving valuable scarce resources and to the important research direction of more (energy) efficient deep learning~\cite{Sze:2017:EPD}.

%% file: sections/supplementary.tex
\section{Overview}
This appendix provides additional explanations and experimental details for reproducibility purposes, which could not be included in the main text due to space limitations.

\section{Addendum to Extension to CNNs}

\begin{figure*}
  \centering
  \includegraphics[width=1.0\linewidth]{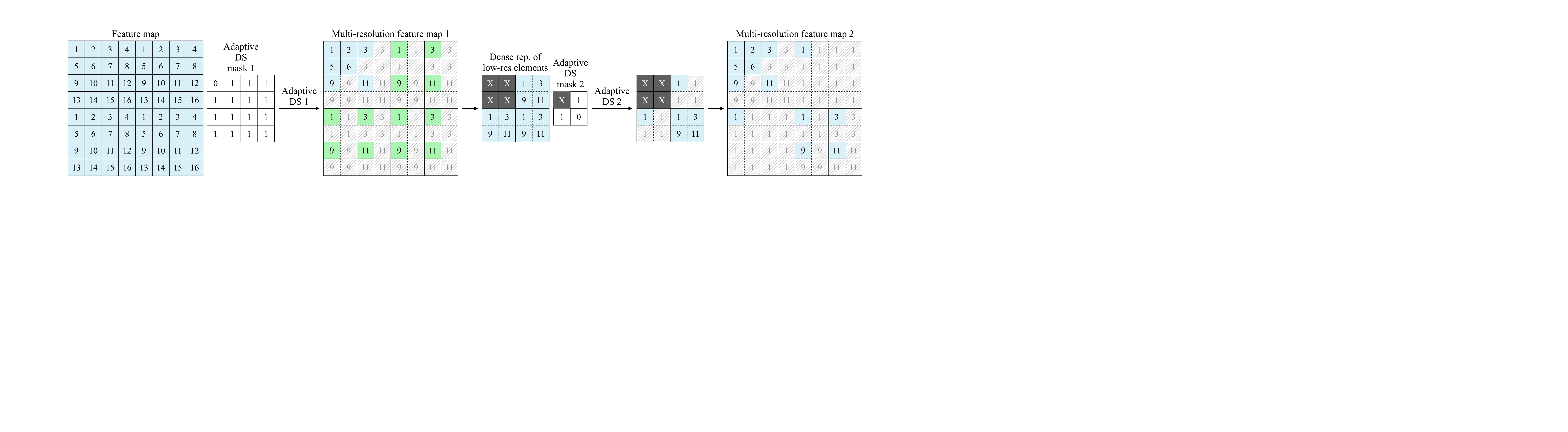}
  \caption[Figure 4]{\emph{Illustration of our adaptive downsampling applied two times in succession with a downsampling factor of $d=2$.} Starting on the left, adaptive downsampling is applied a first time to a regular \emph{feature map} to generate \emph{multi-resolution feature map 1}. To perform adaptive downsampling a second ($k=2$) time, we only consider the $d^k \times d^k$ patches of \emph{multi-resolution feature map 1} that have previously been completely downsampled, \ie, that are of lower effective resolution (green cells). The active elements of these patches are projected back to their regular, dense representation (\emph{dense rep.~of low-res elements}) to apply adaptive downsampling a second time (`X' denotes areas that cannot be represented at the current resolution as they have not been downsampled). The elements of the resulting feature map are then again projected onto their corresponding location to generate \emph{multi-resolution feature map 2}.}
  \label{fig:double}
\end{figure*}

In \cref{sec:methods} of the main paper, we briefly discussed how our method can be applied multiple times in succession to extend it to CNNs. For a more detailed description and illustration of this process, please refer to \cref{fig:double}.

\section{Experimental Details}

In the following, we provide additional details to ensure reproducibility. All experiments have been conducted using Python, PyTorch 1.11.0~\citelatex{Paszke:2017:ADP}, and a single NVIDIA RTX A6000 (48GB) GPU or a single NVIDIA A100-SXM4 (40GB) GPU.

\begin{figure} 
  \centering
  \includegraphics[width=1.0\linewidth]{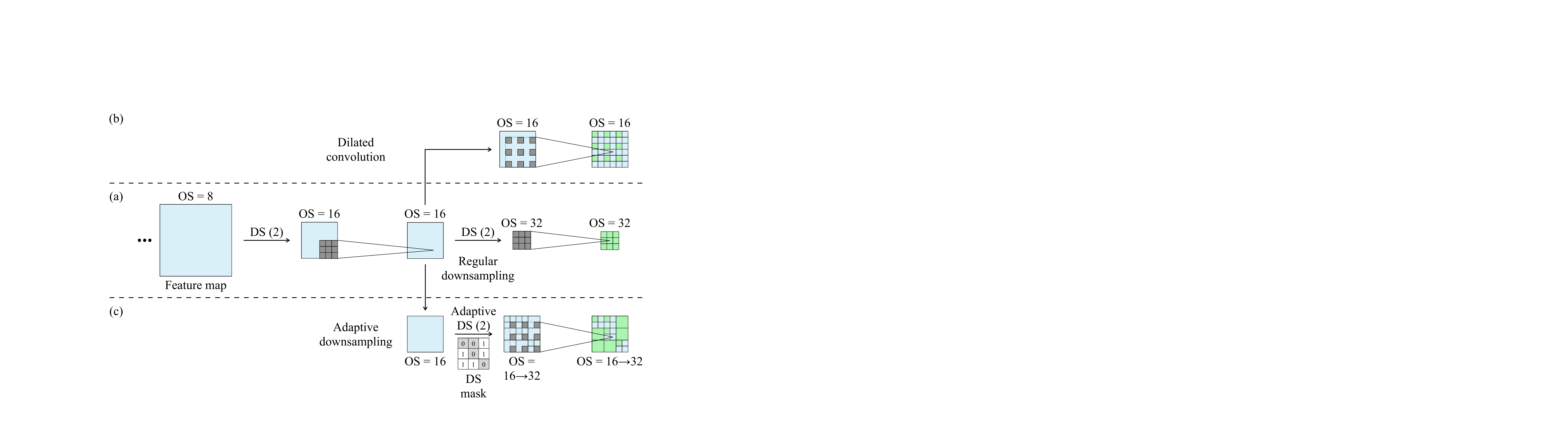}
  \caption[Figure 8]{\emph{Illustration of different methods to handle the output strides (OS) in CNNs.} \emph{(a)} A standard convolutional network progressively downsamples the feature map. The illustrated convolutions, followed by uniform downsampling (DS), correspond conceptionally to strided convolutions with a stride of $2$. \emph{(b)} The last downsampling operation is substituted with a dilated convolution to retain the same receptive field while increasing the feature map resolution~\cite{Yu:2017:DRN}. \emph{(c)} The last downsampling operation is substituted with our novel adaptive downsampling and the dilation is increased to retain the same receptive field. 
  Our approach retains high feature map resolution only where needed.}
  \label{fig:cnn}
\end{figure}

\myparagraph{Implementation of different output strides.}
In all experiments, we examine backbone models with varying output strides.
\cref{fig:cnn} illustrates how these different output strides (OS) are realized. \cref{fig:cnn}\emph{(a)} shows a standard backbone using regular downsampling~\cite{Simonyan:2015:VDC, He:2016:DRL} with the standard downsampling factor $d{=}2$. 
Part \emph{(b)} shows how the backbone can be modified to retain a lower output stride, \ie, higher resolution, by using dilated convolution~\cite{Yu:2017:DRN}. 
Alternatively, it can be modified to contain multiple output strides using our novel adaptive downsampling scheme as shown in \emph{(c)}. 

\subsection{Oracle experiment: Feature resolution in semantic segmentation}\label{sec:app:exp_setup:oracle}

The experiments described in \cref{exp:oracle_seg} of the main paper have been conducted on the widely used Cityscapes dataset~\cite{Cordts:2016:CDS} (freely available to academic and non-academic entities for non-commercial purposes), containing 5\,000 annotated street scene images.
Our code is built on top of the publicly available \texttt{DeepLabV3Plus-Pytorch} GitHub repository (MIT License).\footnote{\href{https://github.com/VainF/DeepLabV3Plus-Pytorch}{\nolinkurl{https://github.com/VainF/DeepLabV3Plus-Pytorch}}}

We examine different ResNet~\cite{He:2016:DRL} backbones (50, 101, and 152 layers), three established segmentation models (FCN~\cite{Long:2015:FCN}, DeepLabv3~\cite{Chen:2017:RAC}, and DeepLabv3+~\cite{Chen:2018:ECA}), and two extensions, \ie Softpool~\cite{Stergiou:2021:RAD} and deformable convolution~\cite{Dai:2017:DCN}. 
Following common practice, we initialize the backbones with weights obtained from pre-training on the ImageNet dataset~\citelatex{Russakovsky:2015:ILS}; they are publicly available in PyTorch. We then fine-tune models with regular output stride on the official Cityscapes~\cite{Cordts:2016:CDS} training split for $30$k iterations using a learning rate of $0.01$ for the backbone, a learning rate of $0.1$ for the classifier, a batch size of $8$, and an SGD optimizer with a weight decay of $1e{-}4$ and a momentum of $0.9$. For the final models with regular and adaptive downsampling, we fine-tune above models for another $10$k iterations with learning rates multiplied by $0.1$. The downsampling mask dilation sizes $K$ for ResNet-\{50, 101, 152\} with DeepLabv3 are $33$, $33$, and $55$, respectively. For ResNet-101 with FCN, respectively DeepLabv3+, $K$ is chosen as $33$. Finally, for ResNet-101 with DeepLabv3 and the extensions deformable convolution and SoftPool, we use $K=55$. SoftPool is used as in the original paper~\cite{Stergiou:2021:RAD}. Deformable convolution is only used in the last block of the ResNet backbone and offsets are estimated using a single convolutional layer with a kernel size of $3 \times 3$ and without a bias.

To measure the theoretical number of multiply-adds, we use the publicly available code in the
\texttt{flops-counter.pytorch} GitHub repository (MIT License).\footnote{\href{https://github.com/sovrasov/flops-counter.pytorch}{\nolinkurl{https://github.com/sovrasov/flops-counter.pytorch}}}
To estimate the runtime in seconds, we use $10$ warup iterations, followed by another $10$ iterations where we measure the average inference time of the backbone.
For a fair comparison between equally optimized implementations, we use the same implementation for our proposed method and the baselines in the main paper. However, in practice, over the years there have been major efforts to highly optimize standard convolution in PyTorch. To put our results into context with these optimizations, in \cref{tab:efficiency} we report inference times for our method, and baseline methods with the default PyTorch implementations. As even the PyTorch version has major impact on the inference time of the baseline models, we report results for two versions, \ie, 1.7.1 and 1.11.0.

Given a reasonable downsampling mask, our proposed method (OS=$8$$\rightarrow$$32$) and implementation can outperform the default PyTorch baseline with an output stride of $8$ (similar mIoU and $0.08$s \vs $0.14$s). For PyTorch 1.7.1, our implementation is even similarly efficient as the default implementation for the regular output strides ($0.08$s for OS=$16$ and $0.26$ \vs $0.24$ for OS=$8$). With version 1.11.0, PyTorch became more optimized, and thus, the gap to our implementation increases for the regular output strides ($0.07$s \vs $0.05$s for OS=$16$ and $0.22$s \vs $0.14$s for OS=$8$). While we are not aware of the nature of the internal optimizations of PyTorch, it is reasonable to assume that similar optimizations could be applied to our implementation.

\begin{table}
  \caption{\textit{Inference times for different implementations.} We consider the ResNet-101+DeepLabv3~\cite{He:2016:DRL, Chen:2017:RAC} setup from the oracle experiment. We report mIoU, theoretical number of multiply-adds (\#MA), and min./max.~inference times for our and the default PyTorch implementation with two different PyTorch versions.}
  \label{tab:efficiency}
  \vspace{-5pt}
  \centering
  \footnotesize
  \begin{tabularx}{\linewidth}{@{}l*5{>{\centering\arraybackslash}X}@{}}
    \toprule
    OS & Impl. & PyTorch version & \#MA & Time(s) & mIoU\\
    \midrule
    $16$ & \textit{Ours} & $1.7.1$ & $4.2e11$  & $0.08$ & $0.7591$\\
    $16$ & \textit{Ours} & $1.11.0$ & $4.2e11$ & $0.07$ & $0.7591$\\
    $16$ & \textit{Default} & $1.7.1$ & $4.2e11$ & $0.08$ & $0.7591$\\
    $16$ & \textit{Default} & $1.11.0$ & $4.2e11$ & $0.05$ & $0.7591$\\
    \midrule
    $8$ & \textit{Ours} & $1.7.1$ & $1.4e12$ & $0.26$ & $0.7775$\\
    $8$ & \textit{Ours} & $1.11.0$ & $1.4e12$ & $0.22$ & $0.7775$\\
    $8$ & \textit{Default} & $1.7.1$ & $1.4e12$ & $0.24$ & $0.7775$\\
    $8$ & \textit{Default} & $1.11.0$ & $1.4e12$ & $0.14$ & $0.7775$\\
    \midrule
    $8$$\rightarrow$$32$ & \textit{Ours} & $1.7.1$ & $6.7e11$ & $0.09/0.18$ & $0.7748$ \\
    $8$$\rightarrow$$32$ & \textit{Ours} & $1.11.0$ & $6.7e11$ & $0.08/0.15$ & $0.7748$\\
    \bottomrule
  \end{tabularx}
\end{table}

\subsection{Case study 1: Semantic segmentation}

The experimental setup in \cref{exp:case_seg} of the main paper follows the same setup as in \cref{sec:app:exp_setup:oracle}. All models are trained from scratch for $30$k iterations. We investigate a ResNet-101~\cite{He:2016:DRL} backbone with the DeepLabv3~\cite{Chen:2017:RAC} segmentation model for OS=16 and OS=8.
The runtime and theoretical number of multiply-adds are also measured as in \cref{sec:app:exp_setup:oracle}. 

The downsampling masks for the edge detection setup are estimated by applying a Sobel filter to the original image, thresholding the result at $0.95$, $0.35$, respectively $0.15$ to control the amount of
invested resources, and dilating the result with a square dilation kernel of size $11 \times 11$.

To estimate the learned downsampling mask, we use a shallow CNN of $4$ layers with a kernel size of $3 \times 3$, with $128$, $64$, $64$, and $2$ channels, with ReLU activation functions, and max pool with a stride of $2$ after the first two layers. The output is fed into a Gumbel-Softmax~\cite{Jang:2017:CRG} to obtain the final discretized downsampling mask. We train the mask estimator with a learning rate of $0.001$ end-to-end together with the segmentation model. The two setups in \cref{fig:seg2} of the main paper are obtained by setting $\gamma$, \ie, the fraction of active elements in the downsampling mask, to $0.5$ and $0.7$.

\subsection{Case study 2: Keypoint description}

For the keypoint description experiment in \cref{exp:case_keypoint} of the main paper, we use the D2-Net descriptor~\cite{Dusmanu:2019:D2N}. Our code is based on the official, publicly available D2-Net~\cite{Dusmanu:2019:D2N} implementation (Clear BSD License).\footnote{\href{https://github.com/mihaidusmanu/d2-net}{\nolinkurl{https://github.com/mihaidusmanu/d2-net}}}
We initialize our models with pre-trained weights that are also available in the official D2-Net~\cite{Dusmanu:2019:D2N} repository and do not train or fine-tune for any of the examined setups.
We estimate the downsampling masks by dilating keypoints obtained from SIFT with filters of varying sizes for each downsampling level. To sample points with different computational complexity in \cref{fig:keypoints_right} of the main paper, we use dilation sizes of --~from least to most multiply-adds~-- ($0$,$0$,$0$), ($0$,$0$,$5$), ($0$,$0$,$11$), ($0$,$0$,$21$), ($0$,$0$,$31$), ($0$,$19$,$35$), ($0$,$27$,$37$), ($19$,$41$,$41$), ($25$,$51$,$51$), ($\infty$,$\infty$,$\infty$). Here, the entries in each triplet correspond to the respective scale they operate on, \ie, the first entry is the dilation size for the mask that retains elements at a resolution of OS${=}1$, the second for retaining elements at a resolution of OS${=}2$, and the third for retaining elements at a resolution of OS${=}4$. An entry of $0$ means that the entire feature map is downsampled while an entry of $\infty$ indicates that the resolution of the entire feature map is retained. As such, ($0$,$0$,$0$) corresponds to the regular output stride of $8$, while ($\infty$,$\infty$,$\infty$) corresponds to the regular output stride of $1$.
The used HPatches dataset~\cite{Balntas:2017:HBE} (MIT License), to the best of our knowledge, contains no personally identifiable information except for one sequence of an image of a prominent person that is already publicly available and not offensive.

%% file: paper.bbl
\begin{thebibliography}{1}\itemsep=-1pt

\bibitem{Paszke:2017:ADP}
Adam Paszke, Sam Gross, Soumith Chintala, Gregory Chanan, Edward Yang, Zachary
  DeVito, Zeming Lin, Alban Desmaison, Luca Antiga, and Adam Lerer.
\newblock Automatic differentiation in {PyTorch}.
\newblock In {\em NIPS Autodiff Workshop}, 2017.

\bibitem{Russakovsky:2015:ILS}
Olga Russakovsky, Jia Deng, Hao Su, Jonathan Krause, Sanjeev Satheesh, Sean Ma,
  Zhiheng Huang, Andrej Karpathy, Aditya Khosla, Michael Bernstein,
  Alexander~C. Berg, and Li Fei-Fei.
\newblock {ImageNet} large scale visual recognition challenge.
\newblock {\em Int. J. Comput. Vision}, 115(13):211--252, 2015.

\end{thebibliography}


\begin{thebibliography}{10}\itemsep=-1pt

\bibitem{Balntas:2017:HBE}
Vassileios Balntas, Karel Lenc, Andrea Vedaldi, and Krystian Mikolajczyk.
\newblock H{P}atches: {A} benchmark and evaluation of handcrafted and learned
  local descriptors.
\newblock In {\em CVPR}, pages 3852--3861, 2017.

\bibitem{Bruckmaier:2020:ACL}
Merit Bruckmaier, Ilias Tachtsidis, Phong Phan, and Nilli Lavie.
\newblock Attention and capacity limits in perception: A cellular metabolism
  account.
\newblock {\em Journal of Neuroscience}, 40(35):6801--6811, 2020.

\bibitem{Chen:2017:RAC}
Liang{-}Chieh Chen, George Papandreou, Florian Schroff, and Hartwig Adam.
\newblock Rethinking atrous convolution for semantic image segmentation.
\newblock {\em arXiv:1706.05587 [cs.CV]}, 2017.

\bibitem{Chen:2018:ECA}
Liang-Chieh Chen, Yukun Zhu, George Papandreou, Florian Schroff, and Hartwig
  Adam.
\newblock Encoder-decoder with atrous separable convolution for semantic image
  segmentation.
\newblock In {\em ECCV}, 2018.

\bibitem{Chen:2023:CFV}
Mengzhao Chen, Mingbao Lin, Ke Li, Yongkui Shen, Wu Yongjian, Fei Chao, and
  Rongrong Ji.
\newblock {CF-ViT}: A general coarse-to-fine method for vision transformer.
\newblock In {\em AAAI}, 2023.

\bibitem{Cordts:2016:CDS}
Marius Cordts, Mohamed Omran, Sebastian Ramos, Timo Scharwächter, Markus
  Enzweiler, Rodrigo Benenson, Uwe Franke, Stefan Roth, and Bernt Schiele.
\newblock The {C}ityscapes dataset for semantic urban scene understanding.
\newblock In {\em CVPR}, pages 3213--3223, 2016.

\bibitem{Dai:2017:DCN}
Jifeng Dai, Haozhi Qi, Yuwen Xiong, Yi Li, Guodong Zhang, Han Hu, and Yichen
  Wei.
\newblock Deformable convolutional networks.
\newblock In {\em ICCV}, pages 764--773, 2017.

\bibitem{Dosovitskiy:2021:IWW}
Alexey Dosovitskiy, Lucas Beyer, Alexander Kolesnikov, Dirk Weissenborn,
  Xiaohua Zhai, Thomas Unterthiner, Mostafa Dehghani, Matthias Minderer, Georg
  Heigold, Sylvain Gelly, Jakob Uszkoreit, and Neil Houlsby.
\newblock An image is worth 16x16 words: Transformers for image recognition at
  scale.
\newblock In {\em ICLR}, 2021.

\bibitem{Dusmanu:2019:D2N}
Mihai Dusmanu, Ignacio Rocco, Tom{\'{a}}s Pajdla, Marc Pollefeys, Josef Sivic,
  Akihiko Torii, and Torsten Sattler.
\newblock D2-{N}et: {A} trainable {CNN} for joint description and detection of
  local features.
\newblock In {\em CVPR}, pages 8092--8101, 2019.

\bibitem{Engelcke:2017:V3D}
Martin Engelcke, Dushyant Rao, Dominic~Zeng Wang, Chi~Hay Tong, and Ingmar
  Posner.
\newblock {Vote3Deep}: Fast object detection in 3{D} point clouds using
  efficient convolutional neural networks.
\newblock In {\em ICRA}, pages 1355--1361, 2017.

\bibitem{Gao:2019:LIP}
Ziteng Gao, Limin Wang, and Gangshan Wu.
\newblock {LIP}: Local importance-based pooling.
\newblock In {\em ICCV}, pages 3354--3363, 2019.

\bibitem{Graham:2014:SSC}
Benjamin Graham.
\newblock Spatially-sparse convolutional neural networks.
\newblock {\em arXiv:1409.6070 [cs.CV]}, 2014.

\bibitem{Graham:2015:S3D}
Ben Graham.
\newblock Sparse {3D} convolutional neural networks.
\newblock In {\em BMVC}, pages 150.1--150.9, 2015.

\bibitem{Graham:2018:3DS}
Benjamin Graham, Martin Engelcke, and Laurens van~der Maaten.
\newblock 3{D} semantic segmentation with submanifold sparse convolutional
  networks.
\newblock In {\em CVPR}, pages 9224--9232, 2018.

\bibitem{Gulcehre:2014:LNP}
Caglar Gulcehre, Kyunghyun Cho, Razvan Pascanu, and Yoshua Bengio.
\newblock Learned-norm pooling for deep feedforward and recurrent neural
  networks.
\newblock In {\em ECML PKDD}, pages 530--546, 2014.

\bibitem{Hariharan:2015:HOS}
Bharath Hariharan, Pablo~Andr{\'{e}}s Arbel{\'{a}}ez, Ross~B. Girshick, and
  Jitendra Malik.
\newblock Hypercolumns for object segmentation and fine-grained localization.
\newblock In {\em CVPR}, pages 447--456, 2015.

\bibitem{He:2016:DRL}
Kaiming He, Xiangyu Zhang, Shaoqing Ren, and Jian Sun.
\newblock Deep residual learning for image recognition.
\newblock In {\em CVPR}, pages 770--778, 2016.

\bibitem{Hesse:2021:FAA}
Robin Hesse, Simone Schaub-Meyer, and Stefan Roth.
\newblock Fast axiomatic attribution for neural networks.
\newblock In {\em NeurIPS*2021}, pages 19513--19524.

\bibitem{Jaderberg:2015:STN}
Max Jaderberg, Karen Simonyan, Andrew Zisserman, and Koray Kavukcuoglu.
\newblock Spatial transformer networks.
\newblock In {\em NIPS*2015}, pages 2017--2025.

\bibitem{Jang:2017:CRG}
Eric Jang, Shixiang Gu, and Ben Poole.
\newblock Categorical reparameterization with gumbel-softmax.
\newblock In {\em ICLR}, 2017.

\bibitem{Jin:2022:LDS}
Chen Jin, Ryutaro Tanno, Thomy Mertzanidou, Eleftheria Panagiotaki, and
  Daniel~C. Alexander.
\newblock Learning to downsample for segmentation of ultra-high resolution
  images.
\newblock In {\em ICLR}, 2020.

\bibitem{Krizhevsky:2012:INC}
Alex Krizhevsky, Ilya Sutskever, and Geoffrey~E. Hinton.
\newblock {ImageNet} classification with deep convolutional neural networks.
\newblock In {\em NIPS*2012}, pages 1106--1114.

\bibitem{LeCun:1989:BAH}
Yann LeCun, Bernhard Boser, John~S. Denker, Donnie Henderson, Richard~E.
  Howard, Wayne Hubbard, and Lawrence~D. Jackel.
\newblock Backpropagation applied to handwritten {ZIP} code recognition.
\newblock {\em Neural Comput.}, 1(4):541--551, 1989.

\bibitem{Lee:2016:GPF}
Chen-Yu Lee, Patrick~W. Gallagher, and Zhuowen Tu.
\newblock Generalizing pooling functions in convolutional neural networks:
  Mixed, gated, and tree.
\newblock In {\em AISTATS}, pages 464--472, 2016.

\bibitem{Lin:2017:FPN}
Tsung{-}Yi Lin, Piotr Doll{\'{a}}r, Ross~B. Girshick, Kaiming He, Bharath
  Hariharan, and Serge~J. Belongie.
\newblock Feature pyramid networks for object detection.
\newblock In {\em CVPR}, pages 936--944, 2017.

\bibitem{Long:2015:FCN}
Jonathan Long, Evan Shelhamer, and Trevor Darrell.
\newblock Fully convolutional networks for semantic segmentation.
\newblock In {\em CVPR}, pages 3431--3440, 2015.

\bibitem{Lowe:2004:DIF}
David~G. Lowe.
\newblock Distinctive image features from scale-invariant keypoints.
\newblock {\em Int. J. Comput. Vision}, 60(2):91--110, 2004.

\bibitem{Luo:2020:ALL}
Zixin Luo, Lei Zhou, Xuyang Bai, Hongkai Chen, Jiahui Zhang, Yao Yao, Shiwei
  Li, Tian Fang, and Long Quan.
\newblock {ASLF}eat: Learning local features of accurate shape and
  localization.
\newblock In {\em CVPR}, pages 6589--6598, 2020.

\bibitem{Marin:2019:ESL}
Dmitrii Marin, Zijian He, Peter Vajda, Priyam Chatterjee, Sam~S. Tsai, Fei
  Yang, and Yuri Boykov.
\newblock Efficient segmentation: Learning downsampling near semantic
  boundaries.
\newblock In {\em ICCV}, pages 2131--2141, 2019.

\bibitem{Pinheiro:2016:LRO}
Pedro~Oliveira Pinheiro, Tsung{-}Yi Lin, Ronan Collobert, and Piotr
  Doll{\'{a}}r.
\newblock Learning to refine object segments.
\newblock In {\em ECCV}, volume~1, pages 75--91, 2016.

\bibitem{Recasens:2018:LZS}
Adri{\`{a}} Recasens, Petr Kellnhofer, Simon Stent, Wojciech Matusik, and
  Antonio Torralba.
\newblock Learning to zoom: {A} saliency-based sampling layer for neural
  networks.
\newblock In {\em ECCV}, volume~9, pages 52--67, 2018.

\bibitem{Revaud:2019:R2D}
J{\'{e}}r{\^{o}}me Revaud, C{\'{e}}sar~Roberto de Souza, Martin Humenberger,
  and Philippe Weinzaepfel.
\newblock {R2D2:} {R}eliable and repeatable detector and descriptor.
\newblock In {\em NeurIPS*2019}, pages 12405--12415.

\bibitem{Riegler:2017:ONL}
Gernot Riegler, Ali~Osman Ulusoy, and Andreas Geiger.
\newblock Oct{N}et: Learning deep 3{D} representations at high resolutions.
\newblock In {\em CVPR}, pages 6620--6629, 2017.

\bibitem{Ronneberger:2015:UNC}
Olaf Ronneberger, Philipp Fischer, and Thomas Brox.
\newblock {U-Net}: Convolutional networks for biomedical image segmentation.
\newblock In {\em MICCAI}, pages 234--241, 2015.

\bibitem{Saeedan:2018:DPP}
Faraz Saeedan, Nicolas Weber, Michael Goesele, and Stefan Roth.
\newblock Detail-preserving pooling in deep networks.
\newblock In {\em CVPR}, pages 9108--9116, 2018.

\bibitem{Simonyan:2014:DIC}
Karen Simonyan, Andrea Vedaldi, and Andrew Zisserman.
\newblock Deep inside convolutional networks: Visualising image classification
  models and saliency maps.
\newblock In {\em ICLR}, 2014.

\bibitem{Simonyan:2015:VDC}
Karen Simonyan and Andrew Zisserman.
\newblock Very deep convolutional networks for large-scale image recognition.
\newblock In {\em ICLR}, 2015.

\bibitem{Stergiou:2021:RAD}
Alexandros Stergiou, Ronald Poppe, and Grigorios Kalliatakis.
\newblock Refining activation downsampling with {SoftPool}.
\newblock In {\em ICCV}, pages 10337--10346, 2021.

\bibitem{Sundararajan:2017:AAD}
Mukund Sundararajan, Ankur Taly, and Qiqi Yan.
\newblock Axiomatic attribution for deep networks.
\newblock In {\em ICML}, pages 3319--3328, 2017.

\bibitem{Sze:2017:EPD}
Vivienne Sze, Yu{-}Hsin Chen, Tien{-}Ju Yang, and Joel~S. Emer.
\newblock Efficient processing of deep neural networks: {A} tutorial and
  survey.
\newblock {\em Proc. {IEEE}}, 105(12):2295--2329, 2017.

\bibitem{Talebi:2021:LRI}
Hossein Talebi and Peyman Milanfar.
\newblock Learning to resize images for computer vision tasks.
\newblock In {\em ICCV}, pages 487--496, 2021.

\bibitem{Uzpak:2020:STK}
Ali Uzpak, Abdelaziz Djelouah, and Simone Schaub{-}Meyer.
\newblock Style transfer for keypoint matching under adverse conditions.
\newblock In {\em 3DV}, pages 1089--1097, 2020.

\bibitem{Wang:2015:DHR}
Jingdong Wang, Ke Sun, Tianheng Cheng, Borui Jiang, Chaorui Deng, Yang Zhao,
  Dong Liu, Yadong Mu, Mingkui Tan, Xinggang Wang, Wenyu Liu, and Bin Xiao.
\newblock Deep high-resolution representation learning for visual recognition.
\newblock {\em {IEEE} Trans. Pattern Anal. Mach. Intell.}, 43(10):3349--3364,
  2021.

\bibitem{Yu:2014:MPC}
Dingjun Yu, Hanli Wang, Peiqiu Chen, and Zhihua Wei.
\newblock Mixed pooling for convolutional neural networks.
\newblock In {\em RSKT}, pages 364--375, 2014.

\bibitem{Yu:2016:MSC}
Fisher Yu and Vladlen Koltun.
\newblock Multi-scale context aggregation by dilated convolutions.
\newblock In {\em ICLR}, 2016.

\bibitem{Yu:2017:DRN}
Fisher Yu, Vladlen Koltun, and Thomas~A. Funkhouser.
\newblock Dilated residual networks.
\newblock In {\em CVPR}, pages 636--644, 2017.

\bibitem{Yu:2018:DLA}
Fisher Yu, Dequan Wang, Evan Shelhamer, and Trevor Darrell.
\newblock Deep layer aggregation.
\newblock In {\em CVPR}, pages 2403--2412, 2018.

\end{thebibliography}
